\documentclass{article}
\usepackage{spconf,amsmath,graphicx}
\usepackage{textcomp}
\usepackage{xcolor}
\usepackage{color}
\usepackage{booktabs}
\usepackage{subfigure}
\usepackage{float}
\usepackage{makecell}
\usepackage{multirow}
\usepackage{bbding}
\usepackage{pifont}
\usepackage[colorlinks,linkcolor=magenta]{hyperref}

\title{TAKING A CLOSER LOOK AT SYNTHESIS: FINE-GRAINED ATTRIBUTE ANALYSIS \\ FOR PERSON RE-IDENTIFICATION}
\name{Suncheng Xiang$^{1*}$, Yuzhuo Fu$^{1}$, Guanjie You$^{2}$, Ting Liu$^{1}$\thanks{$^{*}$ indicates corresponding author. This research was supported by the National Natural Science Foundation of China under Project (Grant No.61977045) and SJTU-SMARCHIT Joint Laboratory of Smart Building.}}
\address{$^{1}$School of Electronic Information and Electrical Engineering, Shanghai Jiao Tong University \\
$^{2}$College of Intelligence Science and Technology, National University of Defense Technology \\
}
\begin{document}
%
\maketitle
\begin{abstract}
Person re-identification (re-ID) plays an important role in applications such as public security and video surveillance. Recently, learning from synthetic data, which benefits from the popularity of synthetic data engine, has achieved remarkable performance. However, in pursuit of high accuracy, researchers in the academic always focus on training with large-scale datasets at a high cost of time and label expenses, while neglect to explore the potential of performing efficient training from millions of synthetic data. To facilitate development in this field, we reviewed the previously developed synthetic dataset GPR and built an improved one (\textit{GPR+}) with larger number of identities and distinguished attributes. Based on it, we quantitatively analyze the influence of dataset attribute on re-ID system. To our best knowledge, we are among the first attempts to explicitly dissect person re-ID from the aspect of attribute on synthetic dataset. This research helps us have a deeper understanding of the fundamental problems in person re-ID, which also provides useful insights for dataset building and future practical usage.
\end{abstract}
\section{Introduction}
\label{sec1}
Person re-ID aims to identify images of the same person from large number of camera views in different places, which has attracted lots of interests and attentions in both academia and industry. Encouraged by the remarkable success of deep learning methods~\cite{li2014deepreid,he2016deep,szegedy2016rethinking,huang2017densely} and the availability of re-ID datasets~\cite{zheng2015scalable,ristani2016performance,wei2018person}, performance of person re-ID has been significantly boosted. Currently, these performance gains come only when a large diversity of training data is available, which is at the price of a large amount of accurate annotations obtained by intensive human labor. Accordingly, real-world applications have to cope with challenges like complex lighting and scene variations, which current real datasets might fail to address~\cite{xiang2020unsupervisedperson}.

\begin{figure}
\centerline{\includegraphics[width=\linewidth]{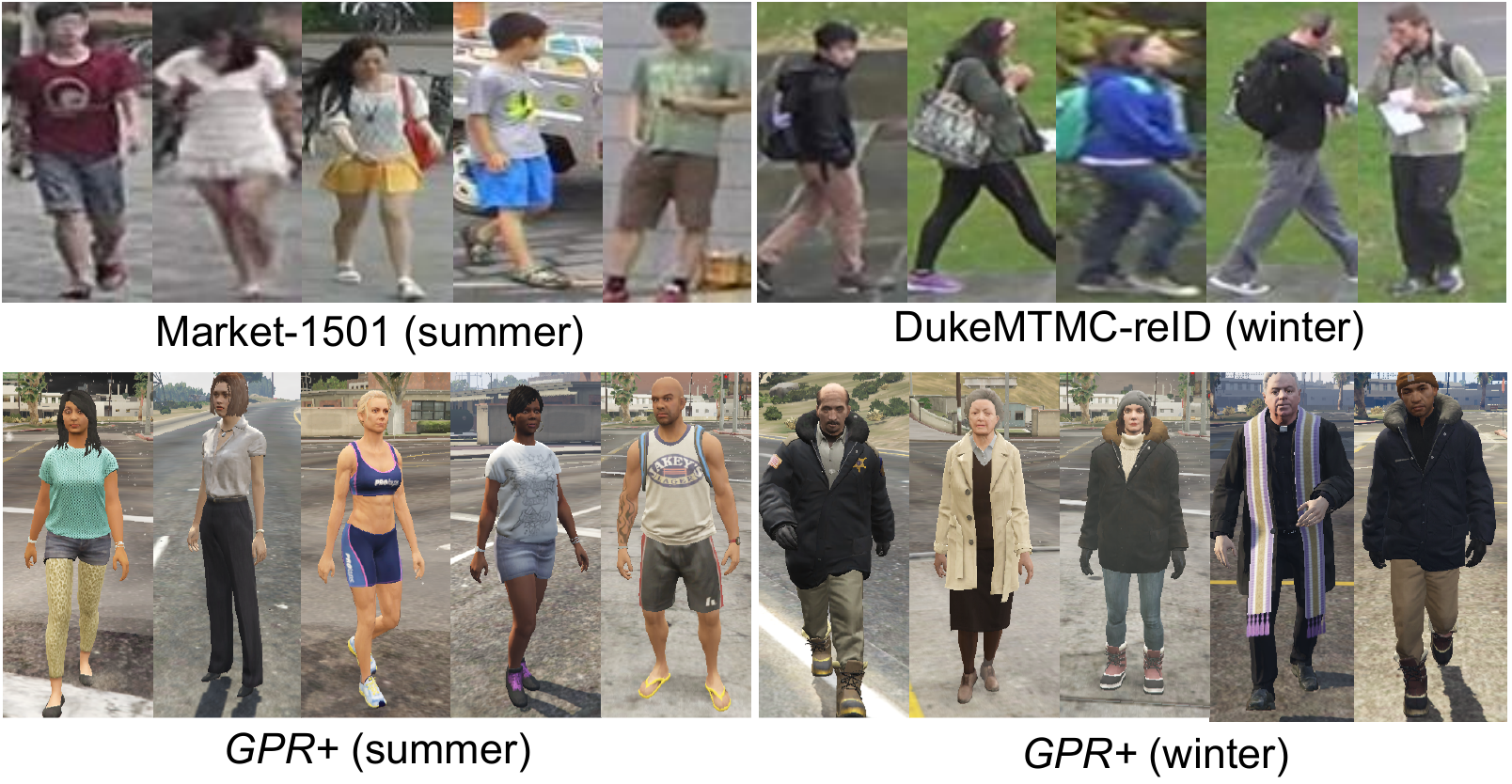}}
\caption{Illustration of the examples between Market-1501 (\textbf{upper-left}) and DukeMTMC-reID (\textbf{upper-right}). It can be easily observed that there exists obvious differences among different datasets in light, background or weather. In comparison, our new dataset \textit{GPR+} (\textbf{bottom}) always has large variances, high resolution and different backgrounds.}
\label{fig1}
\end{figure}

To alleviate this problem, many successful person re-ID approaches~\cite{barbosa2018looking,bak2018domain,sun2019dissecting,wang2020surpassing} have been proposed to take advantage of game engine to construct large-scale synthetic re-ID datasets, which can be used to pre-train or fine-tune CNN network.
For example, Barbosa et al.~\cite{barbosa2018looking} propose a synthetic dataset SOMAset created by photorealistic human body generation software. Sun et al.~\cite{sun2019dissecting} introduce a synthetic data engine to dissect re-ID system on the viewpoints of pedestrian. Recently, Wang et al.~\cite{wang2020surpassing} collect a virtual dataset RandPerson with 3D characters. However, current researches mainly concentrate on achieving satisfactory performance with large-scale data at the sacrifice of expensive time costs and intensive human labors, while neglect the potential of performing efficient training from millions of synthetic data. Besides, in the field of person re-ID, current synthetic datasets mainly provide no more than three attribute annotations, such as IDs, viewpoint or illumination, which cannot satisfy the need for this challenging fine-grained attribute analysis task.

\begin{figure*}
\centerline{\includegraphics[width=\linewidth]{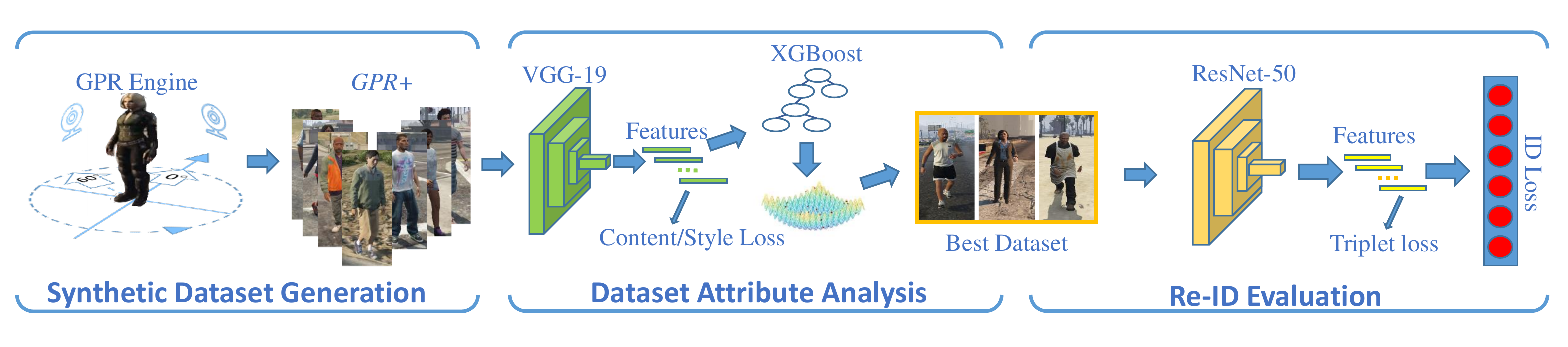}}
\caption{The procedure of our proposed end-to-end systematic framework, which consists of \textbf{1)} synthetic dataset generation (\textit{GPR+}), \textbf{2)} dataset attribute analysis, and \textbf{3)} re-ID evaluation period. Best viewed in color. }
\label{fig2}
\end{figure*}

Another challenge we observe is that, there exists serious scene difference~\cite{yao2019simulating} between synthetic and real dataset. For instance, as shown in Fig.~\ref{fig1}, Market-1501~\cite{zheng2015scalable} only contains scenes that recorded in summer vacation, while DukeMTMC-reID~\cite{ristani2016performance,zheng2017unlabeled} is set in the blizzard scenes. Consequently, pre-training with all synthetic datasets may lead to negative domain adaptation and deteriorate performance on target domain, which is not practical in real-world scenarios.


In this paper, we circumvent above issues by exploring a new direction, that is, analyzing the influences of different attributes in a fine-grained manner. To our knowledge there is no work in the existing literatures that comprehensively study the impacts of multiple attributes on re-ID system. So a natural question then comes to our attention: \emph{how do these attributes influence the retrieval performance? Which one is most critical to our re-ID system?} To answer these questions, we perform rigorous quantification on pedestrian images regarding different attributes. In this paper, our baseline system is built with commonly used loss functions~\cite{hermans2017defense,zhang2018generalized} on vanilla ResNet-50 with no bells and whistles. Moreover, we have summarized a systematic framework for evaluating the importance of different attributes, as shown in Fig.~\ref{fig2}, which can be applied to the construction of datasets with high quality for other computer vision related tasks.

To this end, our work contributes to the research of re-ID community in two different ways: 1) We upgrade the previous dataset GPR to \textit{GPR+}, which is both densely annotated and visually coherent with real world. It also has more identities and decoupled attributes. This is non-trivial,
especially when it comes to the data annotation.
2) We conduct in-depth studies on top of \textit{GPR+} to dissect a person re-ID system comprehensively, then quantitatively analyze the feature importance of different attributes. The empirical results are significant for us to construct high-quality dataset.

\section{Proposed method}
\label{sec2}
\subsection{The \textit{GPR+} dataset}
In previous GPR~\cite{xiang2020unsupervised} dataset, we found that there exists serious redundancy in attribute distribution, for example, the time distribution between ``21$\sim$24" and ``00$\sim$03" is highly correlated that inevitably introduces some interferences in attribute analysis. To address the correlation problem and enhance the orthogonality of intra-attributes, we redefined them in the upgraded version \textit{GPR+}, which provides a solid foundation for fine-grained attribute analysis. In comparison with GPR, we have 1) More identities and bounding boxes; 2) More distinguished and complementary attribute distributions.
We summarize the new features in \textit{GPR+} into the following aspects:

\textbf{Identity.} 808 identities and 475,104 bounding boxes, including more high-quality attribute annotations.

\textbf{Viewpoint.} 12 different types of viewpoint for each identity: every $30^{\circ}$ from $0^{\circ}$ $\sim$ $330^{\circ}$.

\textbf{Weather.} 7 different types of weather: \textit{sunny}, \textit{clouds}, \textit{overcast}, \textit{foggy}, \textit{neutral}, \textit{blizzard}, \textit{snowlight}.

\textbf{Illumination.} 7 different types of illumination: \textit{midnight}, \textit{dawn}, \textit{forenoon}, \textit{noon}, \textit{afternoon}, \textit{dusk}, \textit{night}.

On the whole, these aspects together make \textit{GPR+} a rich dataset for research.
Detailed information and results about \textit{GPR+} can be found at \textcolor{magenta}{\textit{https://JeremyXSC.github.io/GPR/}}.


\begin{figure*}[t]
\begin{minipage}[t]{0.333\linewidth}
\centering
\includegraphics[width=2.52in]{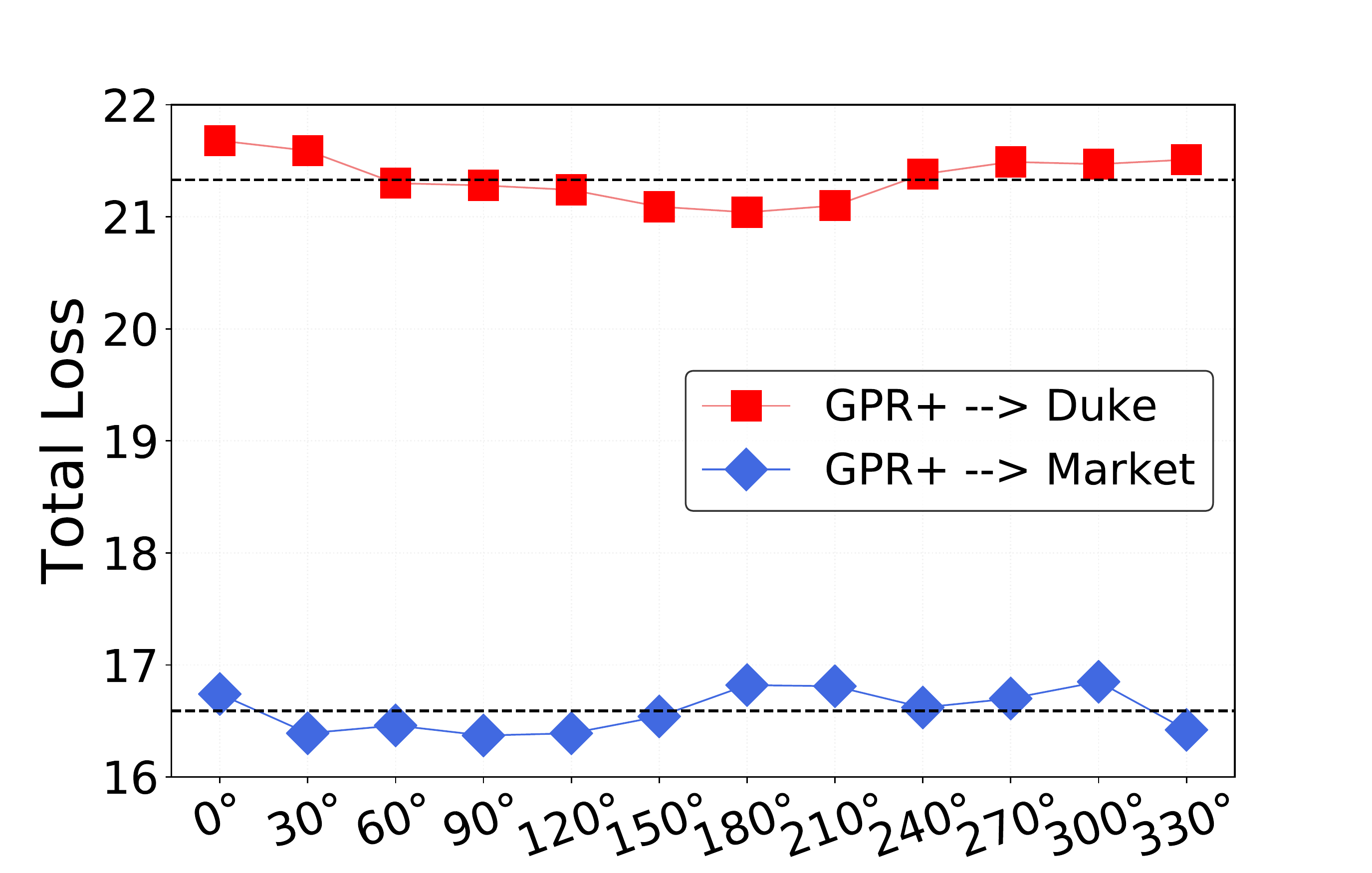}
\centerline{(a) Viewpoint}
\label{figa1}
\end{minipage}%
\begin{minipage}[t]{0.333\linewidth}
\centering
\includegraphics[width=2.52in]{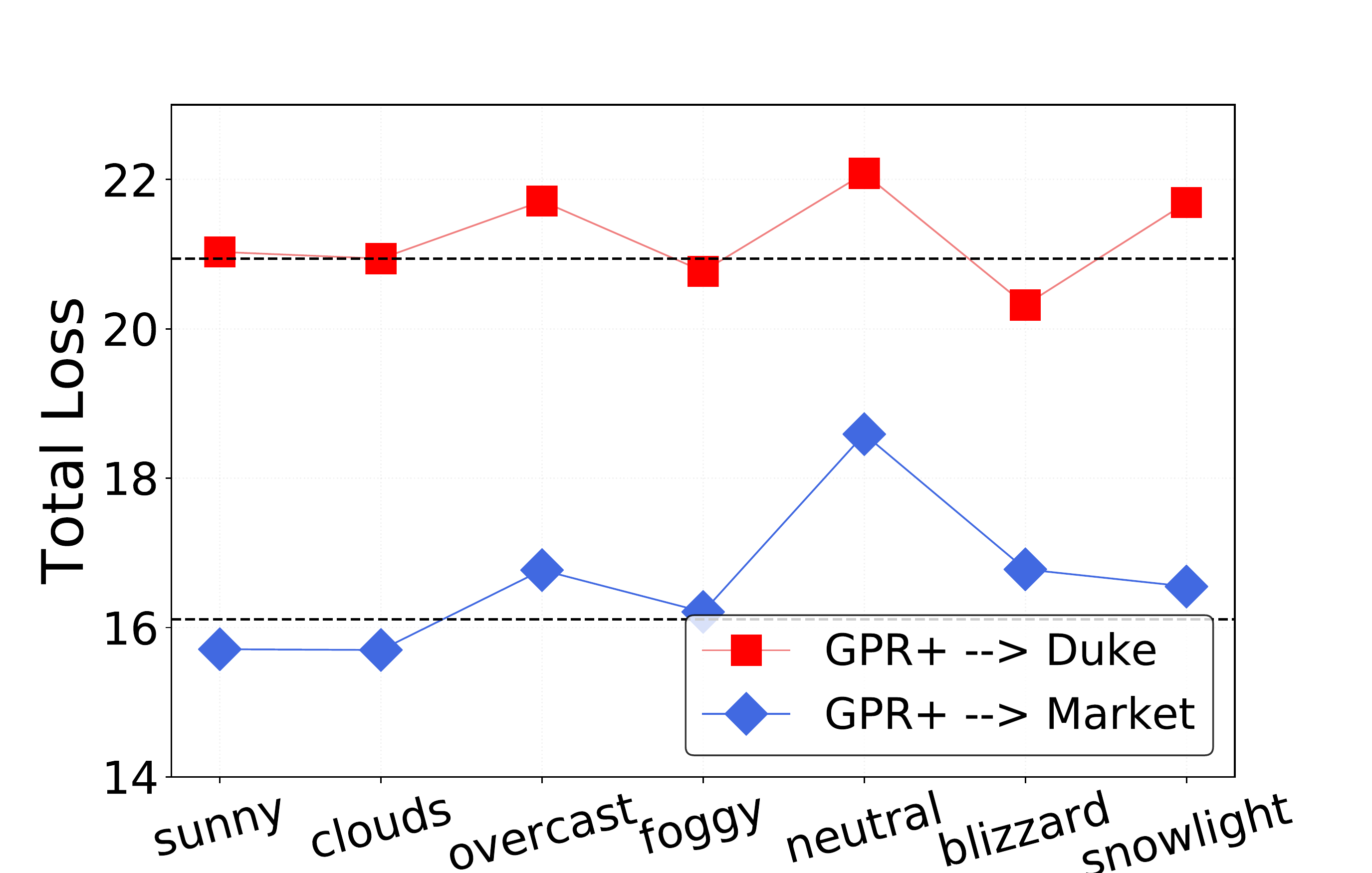}
\centerline{(b) Weather}
\label{figa2}
\end{minipage}
\begin{minipage}[t]{0.333\linewidth}
\centering
\includegraphics[width=2.52in]{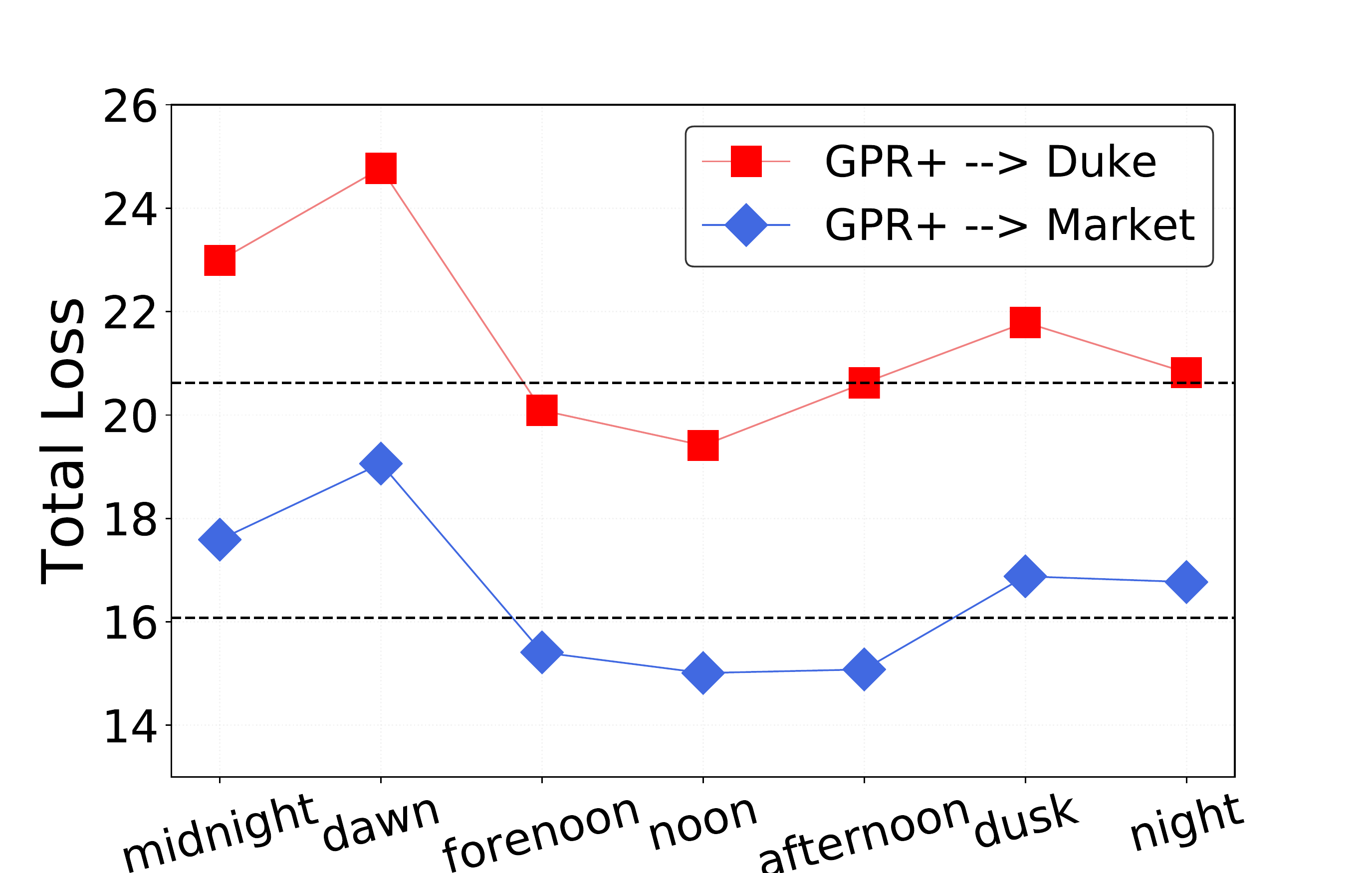}
\centerline{(c) Illumination}
\label{figa1}
\end{minipage}%
\caption{Loss distribution on Duke and Market. Attributes whose value below \underline{horizontal line} can be regarded as optimal items.}
\label{fig4}
\end{figure*}


\subsection{Attribute Analysis}
\label{sec4.1}
One of the key problems in attribute analysis is to find the best representation, which can be addressed from two levels:

\textbf{Content representation.}
In this work, aiming at visualizing the content representation~\cite{gatys2016image} of a image in different layers of CNN, we reconstruct the input image through the feature map of a certain layer in the VGG network~\cite{simonyan2014very}. Under this condition, we adopt squared-error loss between two feature representations $F_{i j}^{l}$ and $P_{i j}^{l}$  of the $i^{th}$ filter at position $j$ in layer $l$ as content loss,
\begin{equation}
\label{eq1}
\mathcal{L}_{\text {content }}=\frac{1}{2} \sum_{i, j}\left(F_{i j}^{l}-P_{i j}^{l}\right)^{2}
\end{equation}

\begin{figure}
\centerline{\includegraphics[width=0.8\linewidth]{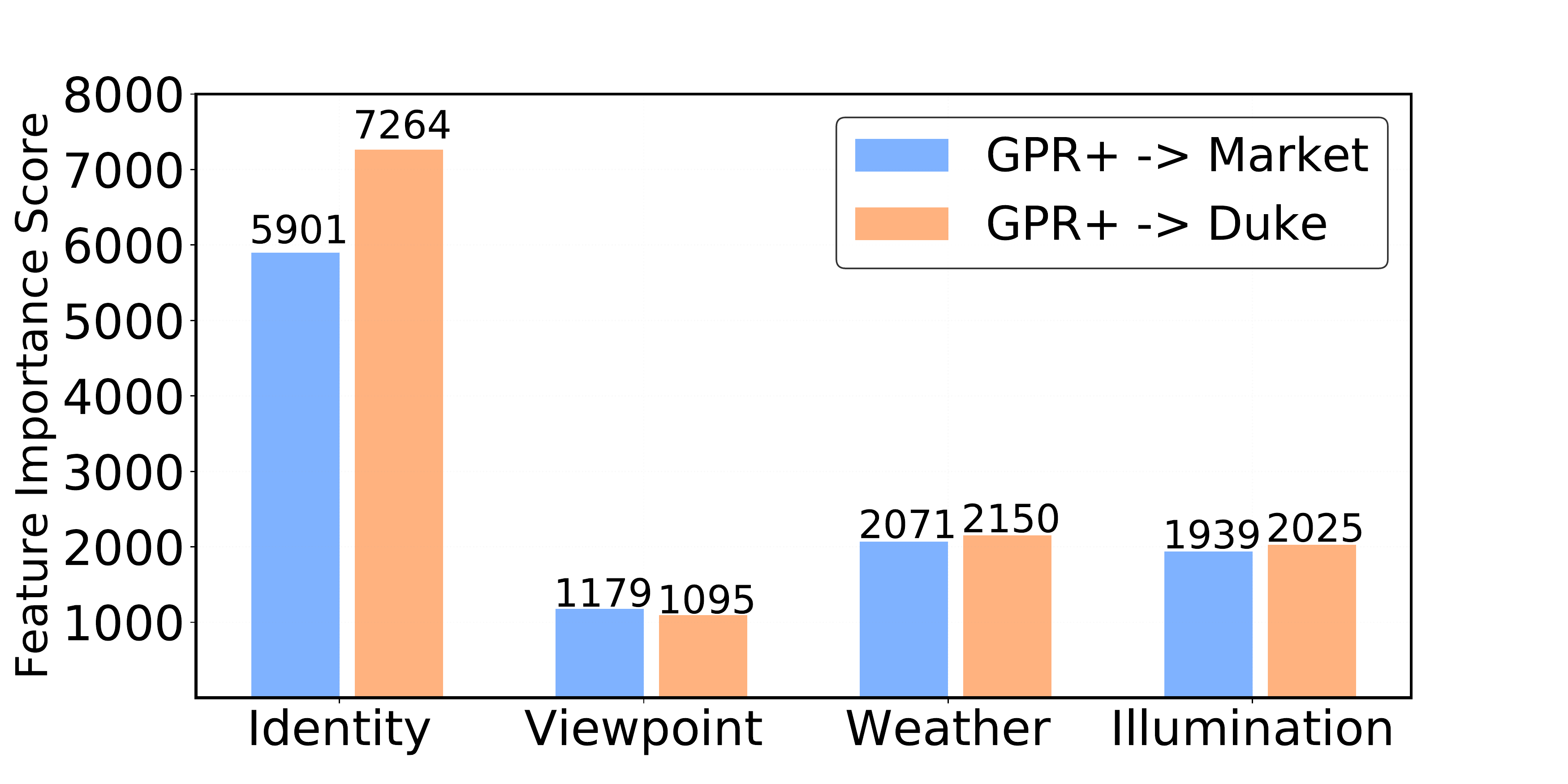}}
\caption{Feature importance of XGBoost with Feature Importance Score (higher is more important).}
\label{fig5}
\end{figure}

\textbf{Style representation.}
Formally, to obtain the style representation~\cite{gatys2017controlling,li2017universal} of input image, we use Gram Matrix to measure the feature correlations between the different filter responses, which is built on top of the CNN responses in each layer of the network.
And the style loss is written as,
\begin{equation}\label{eq2}
\mathcal{L}_{\text {style}}=\sum_{l=0}^{L} w_{l} \frac{1}{4 N_{l}^{2} M_{l}^{2}} \sum_{i, j}\left(G_{i j}^{l}-A_{i j}^{l}\right)^{2}
\end{equation}
where $w_{l}$ is a hyper-parameter that controls the importance of each layer to the style loss $\mathcal{L}_{\text {style}}$. $N_{l}$ represents number of filters and $M_{l}$ is heght $\times$ width of the feature map.
 $G_{i j}$ and $A_{i j}$ denote the Gram Matrix of real images and synthetic images with style representation in layer $l$.
Then the total loss (illustrated in Fig.~\ref{fig4}) for our attribute metric is represented as
\begin{equation}\label{eq3}
\mathcal{L}_{\text {total}} = \alpha * \mathcal{L}_{\text {style}} + \beta * \mathcal{L}_{\text {content}}
\end{equation}
where $\alpha$ and $\beta$ are two hyper-parameters which control the relative importance of style loss and content loss separatively. Then we adopt a scalable end-to-end tree boosting system XGBoost\cite{chen2016xgboost} in terms of \textbf{gain} to analyze the feature importance score of different attributes to re-ID system (see Fig.~\ref{fig5}).



\subsection{Re-ID evaluation}
Re-ID evaluation is the third step of the proposed pipeline, which is same as supervised method. Intuitively, since we mainly focus on \textbf{1) dataset generation} and \textbf{2) attribute analysis}, we follow a widely used open-source$\footnote[1]{\textcolor{black}{https://github.com/Cysu/open-reid}}$ as our standard baseline, and adopt global features provided by backbone ResNet-50~\cite{he2016deep} to perform feature learning. Note that we only modify the output dimension of the last fully-connected layer to the number of training identities. During the period of testing, we extract the 2,048-dim pool-5 vector for retrieval under the commonly used Euclidean distance.

\section{Experimental results}
\label{sec5}
\subsection{Datasets and Implementation Details}
In this paper, we evaluate our method on two benchmark datasets Market-1501~\cite{zheng2015scalable} and  DukeMTMC-reID~\cite{ristani2016performance,zheng2017unlabeled}. Market-1501 has 32,668 person images of 1,501 identities which is split into training set with 751 identities and testing set with 750 identities. All the samples are collected in the campus of Tsinghua University. DukeMTMC-reID contains 1,812 identites. 702 identities are used as the training set and the remaining 1,110 identities as the testing set. For attribute metric, we use the feature space provided by the 19-layer VGG network, and empirically set $w_{l}$=0.2 in Eq.~\ref{eq2}, $\alpha$=0.9 and $\beta$=1 in Eq.~\ref{eq3}. During the re-ID evaluation, we adopt a initializing model pre-trained on ImageNet~\cite{deng2009imagenet} and follow the training procedure in~\cite{luo2019bag,luo2019strong}. All timings for training use one Nvidia Tesla P100 GPU on Pytorch framework~\cite{paszke2019pytorch}.

\begin{table*}[htbp]
  \centering
  \caption{Ablation study results (\%) on Market dataset. \ding{52} indicates using all conditions in each attribute. \textbf{Bold} denotes the best.}
  \normalsize
  \setlength{\tabcolsep}{0.6mm}{
    \begin{tabular}{c c c c c c c c c}
    \Xhline{0.6pt}
    \#identity & \#box   & \#viewpoint & \#weather & \#illumination & Time (h)$\downarrow$   & mAP$\uparrow$  & R@1$\uparrow$  & R@5$\uparrow$ \\
    \hline
    100   & 58,800 & \ding{52}    & \ding{52}     & \ding{52}     & 8.3   & 4.8   & 15.4 &   29.3 \\
    400   & 235,200 & \ding{52}     & \ding{52}     & \ding{52}     & 31.0   & 13.5   & 35.7   & 53.8 \\
    800   & 470,400 & \ding{52}     & \ding{52}     & \ding{52}     & \underline{61.5}   & 17.4  & 41.8   & 60.6 \\
    800    & 134,400 & \ding{52}     & sunny,clouds     & \ding{52}     & 18.0   & \textbf{19.7}   & 43.3   & 59.8 \\
    800   & 201,600 & \ding{52}     & \ding{52}  &  forenoon,noon,afternoon     & 26.5   & 18.6   & 41.4   & 58.8 \\
    800   & 235,200 & $30^{\circ}$, $60^{\circ}$, $90^{\circ}$,$120^{\circ}$,$150^{\circ}$, $330^{\circ}$     & \ding{52}     & \ding{52}     & 30.5   & 19.3   & \textbf{44.1}   & \textbf{62.4} \\
    800   & 28,800 & $30^{\circ}$, $60^{\circ}$, $90^{\circ}$,$120^{\circ}$,$150^{\circ}$, $330^{\circ}$     & sunny,clouds     & forenoon, noon, afternoon  & \textbf{4.5}   & 17.4  & 40.3   & 56.7 \\
    \Xhline{0.6pt}
    \end{tabular}}%
  \label{tab3}%
\end{table*}%

\begin{table*}[htbp]
  \centering
  \caption{Ablation study results (\%) on Duke dataset. \CheckmarkBold indicates using all conditions in each attribute. \textbf{Bold} denotes the best.}
  \normalsize
  \setlength{\tabcolsep}{0.65mm}{
    \begin{tabular}{c c c c c c c c c c}
    \Xhline{0.6pt}
    \#identity & \#box   & \#viewpoint & \#weather & \#illumination & time (h)$\downarrow$   & mAP$\uparrow$   & R@1$\uparrow$   & R@5$\uparrow$\\
    \hline
    100   & 58,800 & \ding{52}     & \ding{52}     & \ding{52}     & 8.3   & 4.3   & 13.8 &   24.2 \\
    400   & 235,200 & \ding{52}     & \ding{52}     & \ding{52}     & 30.6   & 10.7   & 26.3   & 38.2 \\
    800   & 470,400 & \ding{52}     & \ding{52}     & \ding{52}     & \underline{60.7}   & 15.1  & 33.5   & 48.0 \\
    800    & 134,400 & \ding{52}     & foggy,blizzard     & \ding{52}     & 18.0   & 17.8   & 33.8   & 48.3 \\
    800   & 201,600 & \ding{52}     & \ding{52}     & forenoon,noon,afternoon     & 26.5   & \textbf{18.8}   & \textbf{38.2}   & \textbf{52.3} \\
    800   & 235,200 & $60^{\circ}$,$90^{\circ}$,$120^{\circ}$,$150^{\circ}$,$180^{\circ}$, $210^{\circ}$     & \ding{52}     & \ding{52}     & 30.6   & 17.2   & 37.7   & 52.2 \\
    800   & 28,800 & $60^{\circ}$,$90^{\circ}$,$120^{\circ}$,$150^{\circ}$,$180^{\circ}$, $210^{\circ}$     & foggy,blizzard     & forenoon,noon,afternoon       & \textbf{4.4}   & 13.3  & 25.7   & 39.1 \\
    \Xhline{0.6pt}
    \end{tabular}}%
  \label{tab4}%
\end{table*}%


\subsection{Evaluation of Attribute Importance}
\label{sec3.2}
In this section, we evaluate the importance of different attributes on a basic re-ID system. According to Fig.~\ref{fig5}, it can be easily observed that \textbf{\textit{identity}} accounts for the largest proportion when performing cross-domain re-ID task from \textit{GPR+} to Market-1501, following by \textbf{\textit{weather}}, \textbf{\textit{illumination}} and \textbf{\textit{viewpoint}}. Typically, this conclusion is consistent with results on DukeMTMC-reID dataset, which helps us fully understand the role of different attributes.

\subsection{Fine-grained Attribute Analysis}
In this part, we further explore the impacts of different attributes on a basic re-ID system. There are several observations which can be made as follows.

First, we can easily observe that using more IDs will noticeably improve the re-ID performance. For example, as presented in Table~\ref{tab3} and Table~\ref{tab4}, we can only achieve a performance of 4.8\% and 4.3\% in mAP accuracy when tested on Market-1501 and DukeMTMC-reID, respectively. Moreover, adding IDs to 800 as supervised information notably improves the re-ID accuracy, leading to \textbf{+12.6\%} and \textbf{+10.8\%} improvement in mAP accuracy.

Second, we perform greedy search for the objective of \textbf{smaller loss} (see Fig.~\ref{fig4}) to obtain some optimal attributes. For instance, with constraint of 800 IDs, using \emph{sunny\&clouds} and \emph{foggy\&blizzard} bring about +2.3\% and +2.7\% more improvement than using all weather conditions when tested on Market-1501 and DukeMTMC-reID respectively;
using \emph{forenoon, noon, afternoon} as induction can lead an additional improvement of +1.2\% and +3.7\% in mAP accuracy. Furthermore, we can achieve a significant improvement of \textbf{+2.3\%} and \textbf{+4.2\%} in rank-1 accuracy with some critical viewpoints. Intuitively, by taking all attributes into consideration, we can obtain the rank-1 accuracy to \textbf{40.3\%} and \textbf{25.7\%} on Market-1501 and DukeMTMC-reID respectively. Meanwhile, fast training is our second main advantage, e.g., training time will be considerably reduced by $\textbf{13$\times$}$ (61.5 vs. 4.5 hours) and $\textbf{14$\times$}$ (60.7 vs. 4.4 hours), respectively, leading a more computationally efficient training on re-ID backbone.


\subsection{Discussion}
To go even further, we gave an explanation about several interesting phenomenons observed during the experiment.

Firstly, as depicted in Table~\ref{tab3} and Table~\ref{tab4}, it can be observed easily that with only one individual attribute constraint can obtain a more satisfactory performance compared with using all images, no matter which attribute is adopted. It probably because that our attribute analysis strategy can find important attributes and reduce the redundancy of training set.

Secondly, a simply combination of several attributes cannot always guarantee the most optimal attributes for re-ID task, and the mutual influences of multiple attributes should be considered for fine-grained attribute analysis in the future.

Third and importantly, by simply increasing the scale of training examples does not automatically bring notable performance gains. However, using more IDs as training samples is always beneficial to the system. Based on this observation, we can drastically improve our performance by enhancing the diversity of training set instead of increasing the scale of dataset.
In summary, our fine-grained attribute analysis strategy is similar to~\cite{yao2019simulating}, but our solution is entirely parameter free, which makes it more flexible and adaptable.


\section{Conclusion}
\label{sec4}
In this paper, we addressed a critically important problem in person re-identification which has received litter attention thus far - fine-grained attribute analysis. First, we upgrade and enrich the previous GPR dataset to \textit{GPR+}, which provides orthogonal distribution and eliminates the correlation between different attributes. Based on \textit{GPR+}, we introduce a fine-grained analysis strategy to quantitatively assess the importance of attributes, then conduct comprehensive experiments to explore the influences of various attributes on re-ID task. Nevertheless, this research is very meaningful since it will provide guidance for us to construct a high-quality re-ID dataset. In closing, we hope that our new experimental evidence about \textit{GPR+} and it role will shed light into potential future directions for the community to move forward.


\bibliographystyle{IEEEbib}
\bibliography{refs}

\end{document}